# Fractal Gripper: Adaptive manipulator with mode switching

Jiaxin Huang, Jian Shen, Yilin Zheng, Zhigong Song*
School of Mechanical Engineering, Jiangnan University, Wuxi 214122, China.
song_jnu@jiangnan.edu.cn

**Abstract:** Although the multi-jointed underactuated manipulator is highly dexterous, its grasping capacity does not match that of the parallel jaw gripper. This work introduces a fractal gripper to enhance the grasping capacity of multi-joint underactuated manipulators, preserving their passive clamping features. We describe in detail the working principle and manufacturing process of the fractal gripper. This work, inspired by the 'Fractal Vise' structure, resulted in the invention of a fractal gripper with mode switching capabilities. The fractal gripper inherits the inherent adaptive properties of the fractal structure and realizes the self-resetting function by integrating spring into the original design, thereby enhancing the efficiency of object grasping tasks. The fractal gripper prevents object damage by distributing pressure evenly and applying it at multiple points through its fractal structure during closure. Objects of various shapes are effectively grasped by the fractal gripper, which ensures a safe and secure grasp. The superior performance was provided by the force distribution characteristics of the fractal gripper. By applying the flexible polymer PDMS, which possesses superior elasticity, to the fractal structure's wrapping surface, potential scratching during grasping is effectively prevented, thus protecting the object's geometric surface. Grab experiments with objects of diverse shapes and sizes confirm fractal gripper multi-scale adaptability and superior grasping stability.

## 1 INTRODUCTION

Manipulators, as key components of robotic actuators, have been long recognized in the field of spatial grasping and object manipulation as a research hotspot [1-3]. Incorporating embodied intelligence techniques [4] allows researchers to more accurately mimic the perception and response mechanisms of biological systems, making manipulators smarter and more adaptive. Popular artificial intelligence methods [5-7], known for outstanding feature learning and pattern recognition, enhance manipulator perception, control, and decision-making abilities. Deep neural networks, combined with vision [8,9], force [10], and haptic [11,12] perception, are used in current mainstream robotic end-grasping methods. However, the sensory components and computational costs required by this approach are considered to be too large [13]. The quality of grasping [14,15] is identified as a major factor influencing the performance and reliability of a manipulator. A wide range of aspects, including the stability, accuracy, safety, and adaptability of the manipulator to different shapes, sizes, and materials of the target object, are encompassed by grasping quality. Hence, the understanding and optimization of grasping quality are deemed crucial for the enhancement of the efficiency and usefulness of a manipulator.

Currently, anthropomorphic [16,17] dexterous hands [18] represent a significant research area in robotics. Notably, the ILDA [19], an anthropomorphic high dexterity manipulator featuring an integrated linkage drive, 15 degrees of freedom across 20 joints, a weight of 1.1 kg, and tactile sensing capabilities, stands out. However, its control system presents challenges due to its high dexterity. In addition, flexible soft grippers [20-25], known for their flexible and adaptive wrapping characteristics, are widely recognized and utilized in robotic gripping. General soft grippers are designed to be lightweight, reducing overall system weight, but may not provide enough force for grasping tasks involving heavier objects. The planar linkage gripper [26], the most widely used gripping solution today, typically features a simple two-finger or multi-finger design suitable for general tasks. However, it falls short in adaptability, especially when handling irregularly shaped objects. To address the limitations of planar gripping, underactuated [27,28] and multi-jointed manipulators [29,30] have emerged, typically in three-finger [29,31] or four-finger [32,33] configurations. Multi-joint manipulators with mode switching [31,33] have shown greater adaptability for various object grasping tasks. However, these synergistic hand designs may be confined to specific sets of grasping tasks. Subsequently, the emergence of many novel (special-purpose) grippers, including miniature grippers [34,35], reconfigurable grippers [36], tendril-like grippers [37], trap-like grippers [38], jellyfish-like grippers [39], entangling grippers [40], twist grippers [41], origami grippers [42-44], resilient seashell grippers [45], suction cup grippers [46], inanimate spider grippers [47], and bionic bird grippers for drone landing [48], has been observed. These uniquely shaped gripping solutions necessitate serious consideration regarding their maintenance.

In 1913, a patent was granted to Paulin Karl Kunze for 'Device for obtaining intimate contact with, engaging, or clamping bodies of any shape' [49]. Filed in 1912, the applications encompassed by this patent continue to be widely utilized in the field of robotics today. The impact of the uses covered by this patent extends significantly, influencing not only the research community but also the industrial and commercial sectors [13]. The patent was subsequently named 'Fractal Vise' and initially found a niche market in mechanic stores. Over time, the 'Fractal Vise' began to gain recognition and demonstrated broad potential and applicability. Unfortunately, despite the potential of Kunze's concept and vision, the manufacturing limitations of that era hindered the device's performance and popularity. Material selection, process technology, accuracy limitations, and part availability, all factors in manufacturing constraints, are known to affect device performance and reliability adversely. With the emergence of 3D printing technology and the rapid advancements in manufacturing processes, significant potential and applicability have been demonstrated by fractal structures in the modern era, attributed to additive manufacturing. The flexibility of this technology and the advantages of the manufacturing process have created favorable conditions for the implementation of fractal structures, opening up new opportunities for innovation and progress in various fields. The innovative concepts and technologies revealed in this patent have the potential to make an important contribution to the field of robotics, both then and now. Despite the rapid advances in robotics over the past century, the designs in the patent have remained universal, providing key insights and theoretical foundations for today's robotic systems. By inheriting and applying the fractal structure from the patent, modern robots are enabled to operate more efficiently, accurately, and safely, offering valuable guidance for further exploration and innovation in future robotics.

In this work, a fractal gripper with mode switching is proposed, which achieves excellent spatially stable grasping in different modes (grasping and gripping). A fractal gripper, a type of manipulator, is designed based on the principles of fractal geometry [50]. Leveraging the principle of minimum potential energy, it redistributes force such that the smaller the part of the fractal, and the lower the strength of the fractal structure of the part, the more complex the overall structure becomes. During the closing action of the fractal gripper's end contacts, each sector, being able to rotate around its local coordinate system, ensures that as long as any tentacles have not fully contacted the object to be grasped, a non-zero moment will persist, causing rotation until full contact is achieved. The mechanical effect of the fractal structure is consistent at every level. Regardless of the complexity of the object's shape, when all tentacles are in full contact with the object, pressure is distributed more completely, creating sufficient and balanced friction to maintain the object in place. During closure, the mechanical advantage of the fractal gripper lies in distributing pressure over as large a surface area and as many contact points as possible, thereby preventing inappropriate localized stresses from deforming or damaging the contour of the gripped object. The design of the fractal gripper enables it to clamp objects of various shapes more efficiently, offering a more stable and reliable grasping capacity. The force distribution characteristics of the fractal gripper ensure superior performance during grasping tasks. This mechanical advantage is critical in various applications, such as robotics, automated operations, and agricultural harvesting, facilitating efficient, precise, and safe grasping. A fractal gripper, characterized by its force distribution, is also multi-scale adaptive, enabling it to accommodate objects of various sizes and shapes without extensive structural changes. This adaptability proves valuable in applications where objects of different sizes need to be grasped and manipulated. The principles of fractal geometry have been applied in the design of seismic structures [51] and shock-absorbing structures [52]. Similar to fractal grippers are Whiffletrees [53,54], which also feature self-adjustment and force distribution and show promising applications. Grippers utilizing particulate clogging [55,56] and multi-scale universal grippers [57] are equally adaptive to various object shapes, but they lack the structural stiffness of fractal grippers with rigid fractal structures.

The self-similarity feature of the fractal gripper allows for the reapplication of similar design principles at different scales, simplifying manufacturing and reducing costs. Additionally, the wear-resistant (redundant) nature of fractal grippers ensures that even if one part is damaged, the other parts continue to function properly, enhancing the system's reliability and durability. Kunze's original patent for a 'Fractal Vise' is characterized by having only one degree of freedom for actuation, relying on a screw drive mechanism for clamping, and featuring $2^{n+1}-1$ joints, where ' n ' is the design parameter defining the fractal structure, analogous to the structural depth of a binary tree. A two-dimensional sketch illustrating the fractal structure in a fractal gripper with a grasping case is presented (Fig. 1).

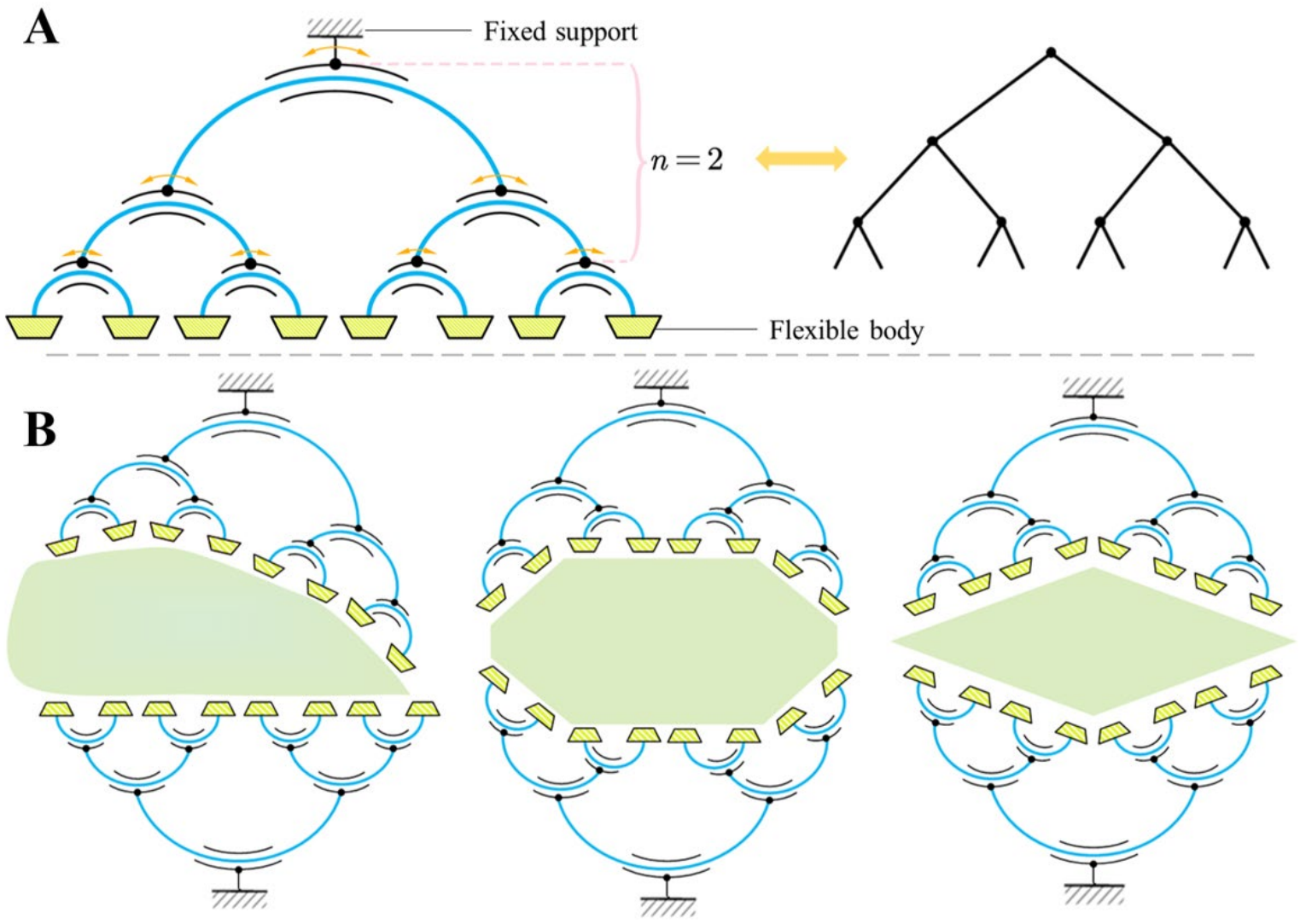

**Fig.1. Two-dimensional schematic and planar fixed case of fractal structure.** (**A**) A simplified two-dimensional planar diagram of a fractal structure, corresponding to a binary tree structure with a depth of 2, possesses good adaptability and force distribution characteristics. (**B**) For objects of different geometric scales, three cases of planar gripping using fractal structures are presented. Relying on the adaptive characteristics of the fractal structures, the geometric surface of the objects can be fully enveloped.

The original fractal vise design, featuring a sliding rotating dovetail joint with two movable jaws controlled by a screw, divided each jaw into four tiers. Each tier, in turn, possessed a similar structure and function, with sectors and partial rotation. Rotation was facilitated by sliding slots with finite holes, ensuring that each sector could rotate back and forth within a certain angle. This design allowed the last tier of contacts to adapt to the complex shape of the workpiece, with each sector joint being capable of being actuated or submissive. The return of the fractal structure to its initial position after object removal can be achieved using springs or magnets. Possible applications include clamping frames (such as pliers and lathe chucks), clamping devices (like limb and head clamps, and foot clamps for surgical and similar instruments), support devices (circular bearings, brackets, chair seats and backrests, surgical and handicapped chairs), sheet metal straightening machines, among other uses [49]. In Kunze's 'Fractal Vise', a fractal structure design parameter of $n = 3$, results in a total of 15 joints, illustrating that with an increase in fractals, parts become smaller, and the strength of the fractal structure part decreases, thereby increasing the complexity of the overall structure. For the fractal gripper, the fractal structure design parameter is set at $n = 2$, resulting in a total of 7 joints. The lightweight of the gripper is considered crucial. Topology optimization [58,59] can be employed for this purpose, or artificial holes can be introduced into the fractal structure to reduce mass, with the stipulation that the mechanical properties of the gripper and gripping strength must not be excessively compromised. For the design reference source of the fractal structure described in this work and the overall 3D structure of the fractal gripper to the physical visualization see (fig. S1).

## 2 RESULTS

**Structural design of fractal gripper and drive mechanism**

This section outlines the structural design of the fractal gripper, where the fractal structure is incorporated into the grasping mechanism, resulting in a manipulator with adaptive capability. In this work, a modified version of the 'Fractal Vise' structure is presented, along with the proposal of a manipulator designed for spatial adaptive grasping, thereby expanding the application beyond the original planar gripping. The fractal gripper's features inherit the adaptive function of the original fractal structure, and with the addition of springs, a self-reset function is achieved. Compared to the existing bionic flexible adaptive wrapping structures on the market, the fractal gripper designed in this work is expected to have superior grasping strength and structural rigidity. The fractal structure adopts a half-and-half design, where each part is divided into two halves and assembled using screws and nuts. Differing from the 'Fractal Vise' mating structure, the original dovetail groove is replaced with a T-slot, and a restraining structure on the edges of the scalloped part prevents it from rotating out of the T-slot. The orange circle marks the screw hole's position, while the blue circle indicates the fixed roll pin's location, which is utilized to achieve the end fixation of the spring. The spring facilitates the self-resetting effect of the fan-shaped parts between each T-shaped slot (in Fig. 2C). Additionally, the flexible body attached to the fractal contact surface not only protects the surface of the object to be grasped but also inherits the high strength of the rigid structure, effectively safeguarding the object's geometric surface during the grasping process. The new robot end gripper (fractal gripper) proposed in this work holds significant potential for re-optimizing the design space. The overall structural design of the fractal gripper is depicted (Fig. 2).

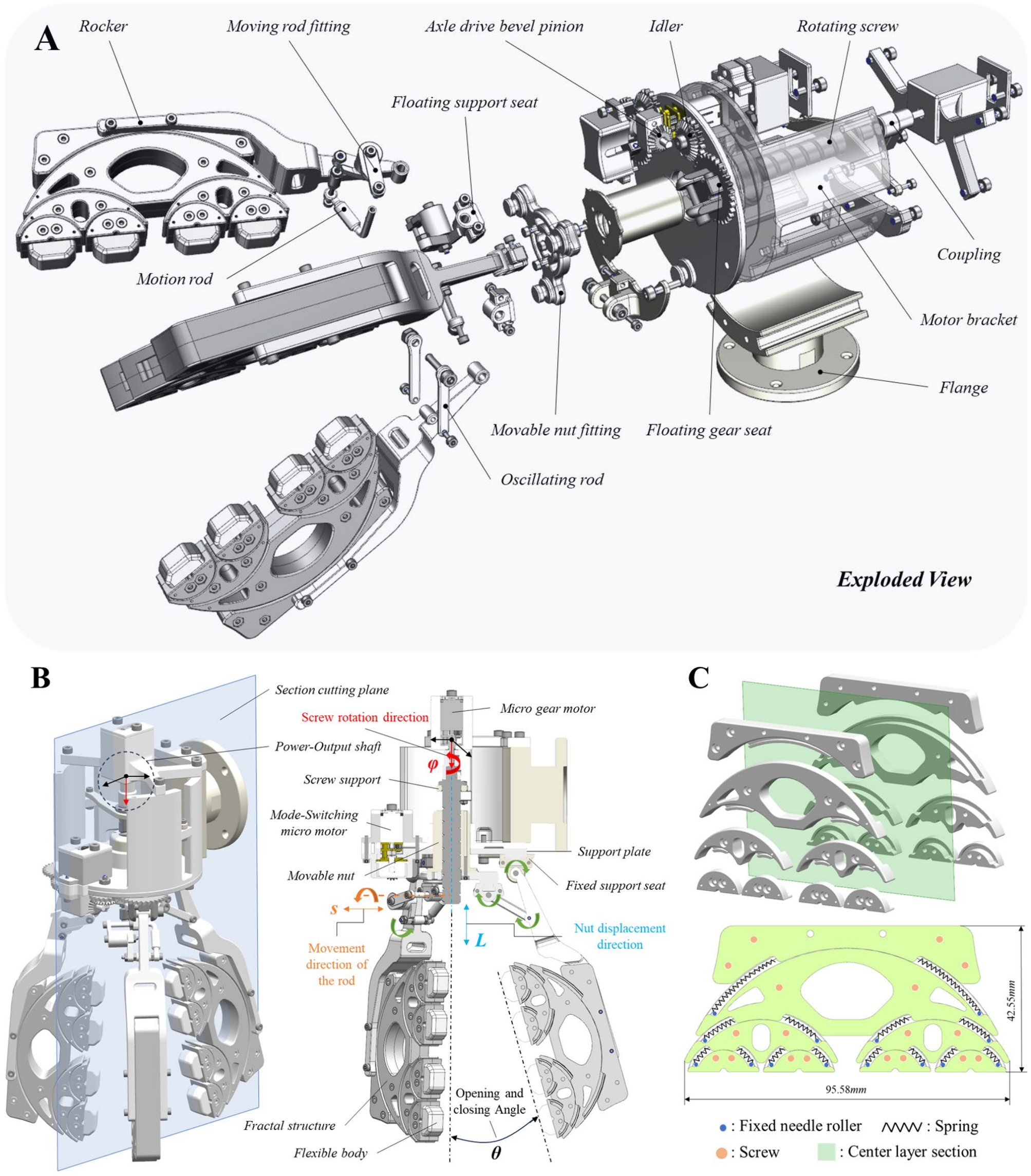

**Fig.2. The overall structure design of the fractal gripper.** (**A**) An exploded view of the overall structure of the fractal gripper. (**B**) A schematic diagram of the driving mechanism of the fractal gripper, which achieves the opening and closing of three fractal structures through spiral transmission and has a self-locking feature. (**C**) A half-and-half design diagram of the fractal structure, dividing each part into two halves, assembled using screws and nuts. Between each T-slot, a spring is used to achieve the self-resetting effect of the fan-shaped parts.

In the structural design of the fractal gripper, a single drive motor and fractal structure are utilized to grab almost any irregularly shaped spatial object safely. For successful grasping, the object must conform to the design space defined by the fractal gripper. In the fractal gripper's grasping device, a micro gear motor is employed for the actuation of the robot hand's opening and closing. Additionally, the adjustment of grasping and gripping attitudes is facilitated by a worm-and-gear biaxial microform motor for mode switching. For the actuation of the fractal gripper's closing and opening actions, a linkage scheme involving a micro gear motor is employed. This motor drives a rotating screw, working in conjunction with a movable nut and linkage, to achieve an adaptive grasping effect on the grasped object. On the grasping surface of the fractal structure, a flexible polymer, specifically PDMS [60], is applied. The elasticity of the PDMS structure helps to avoid potential scratching by the manipulator during grasping, effectively protecting the object's geometric surface. This PDMS structure is referred to as the flexible body in this work. Other details about the fabrication method of the flexible body and the fractal gripper can be seen (fig. S2).

The drive mechanism part is designed as a single degree of freedom reciprocating movement, utilizing a rocker in conjunction with a screw to achieve a fixed constraint on the fractal structure. The bottom end of the rocker, relying on the screw and a locking nut, achieves rotational cooperation with a floating gear seat on the support plate. This setup enables mode switching via the floating gear seat. The rotation of the rotating screw, driven by a micro gear motor, facilitates the linear movement of the movable nut and its accessories. The use of a screw drive achieves the linkage effect between the rocker and the fractal structure. Lastly, the end of the rotating screw, along with the coupling, is secured by a locking nut. A hole is centrally located in the support plate and the motor bracket, forming a clearance fit with the movable nut. The micro gear motor and screw support are installed on the motor bracket. The screw support's role is to ensure the axial running of the rotating screw under excessive clamping force during movement and to prevent its detachment from the coupling. A round hole is provided in the middle section of the rocker to facilitate linkage with the oscillating rod. The other end of the oscillating rod rotates with the floating support seat on the movable nut fitting, allowing the floating support seat to rotate around the movable nut fitting. Additionally, holes are provided on the rear face of the motor bracket to accommodate an external vision module. A flange is added to the lower end of the motor bracket, which is integrated with both the motor bracket and the support plate. This integration, facilitated by the flange, allows the fractal gripper to be attached to the robotic arm. For different applications, the parts of the fractal gripper can undergo parametric design, ensuring the entire grasping system's stability and reliability. The modular design facilitates easier system integration of the fractal gripper. The screw drive mechanism possesses a self-locking characteristic, achieved when the helical rise angle is less than the equivalent friction angle between the teeth. This characteristic enables self-locking in the event of a power failure, ensuring that the gripper remains secure after the object has been grasped and the motor has stopped, preventing slackening due to the object's excessive weight.

The direct drive of the screw drive can be achieved through the use of a micro gear motor and a coupling, with the number of motor revolutions being recorded as $\varphi$. The stroke $L$ required by the movable nut for the transition from closing to opening of the fractal gripper is established at 15mm. The opening angle $\theta$ of the rocker, defined as the angle between the center axis and the rotating screw, has its maximum value recorded as $\theta_{max} \approx 32°$. The performance parameters for the micro gear motor, utilized in both the drive mechanism (screw drive) and the mode switching part, are presented in Table 1.

**Table 1. The performance parameters of micro gear motor.**

| Micro gear motor | Screw drive | Mode-Switching |
|---|---|---|
| Output shaft | Uniaxial | Biaxial |
| Reduction ratio | 298 | 1030 |
| Voltage（V） | 12 | 12 |
| No-load velocity（r/min） | 60 | 16 |
| Load velocity（r/min） | 40 | 14 |
| Rated torque（N·m） | 0.176 | 0.097 |
| Gridlock torque（N·m） | 1.07 | 0.32 |

In the screw drive component of the drive mechanism, a single shaft output micro gear motor is employed, with a load velocity $n_{load}$ = 40 r/min. The rotating screw, being a single-threaded screw ($Z$ = 1), has a pitch ($P$) set at 7mm. The linear displacement velocity ($V$) of the moving nut is as follows, see formula (1).

$$V = n_{load} \cdot P \cdot Z \tag{1}$$

It can be obtained that $V$ = 4.67 mm/s. Continue to give the formula for the number of revolutions ($\varphi$) required to achieve $L$ = 15mm displacement of the nut as follows, see formula (2).

$$\varphi = \frac{L}{P \cdot Z} \tag{2}$$

It can be obtained that $\varphi$ = 2.14 r. The opening and closing velocity of the fractal gripper, defined by the linear displacement velocity of the movable nut, is a critical physical parameter influencing the velocity of the object's grasping action. With a fixed motor velocity, opting for a rotating screw with a larger pitch is known to enhance the movable nut's linear displacement velocity. However, it is also necessary to ensure that the screw drive maintains self-locking performance. Self-locking is characterized by the screw drive's ability to prevent uncontrolled backward or reverse motion in the absence of an external force. To meet the self-locking requirement, the pitch of the screw drive must be set to be less than the friction angle, which is defined as the minimum angle at which the contact surfaces between materials can overcome static friction. In the design of a screw drive mechanism, careful consideration must be given to the choice of pitch. Selecting a larger pitch is known to increase the linear displacement velocity of the fractal gripper, yet ensuring the maintenance of self-locking properties is essential. Through the consideration of pitch, friction angle, and design requirements, the appropriate configuration of parameters can be selected, achieving the desired opening and closing velocity while ensuring the stability and reliability of the gripper.

## Mode switching of the fractal gripper

The mode switching mechanism is primarily comprised of a worm-and-gear biaxial microform motor, axle drive bevel pinion, idler, floating gear seat, floating support seat, motion rod, and moving rod fitting. The effect of mode switching is depicted (Fig. 3).

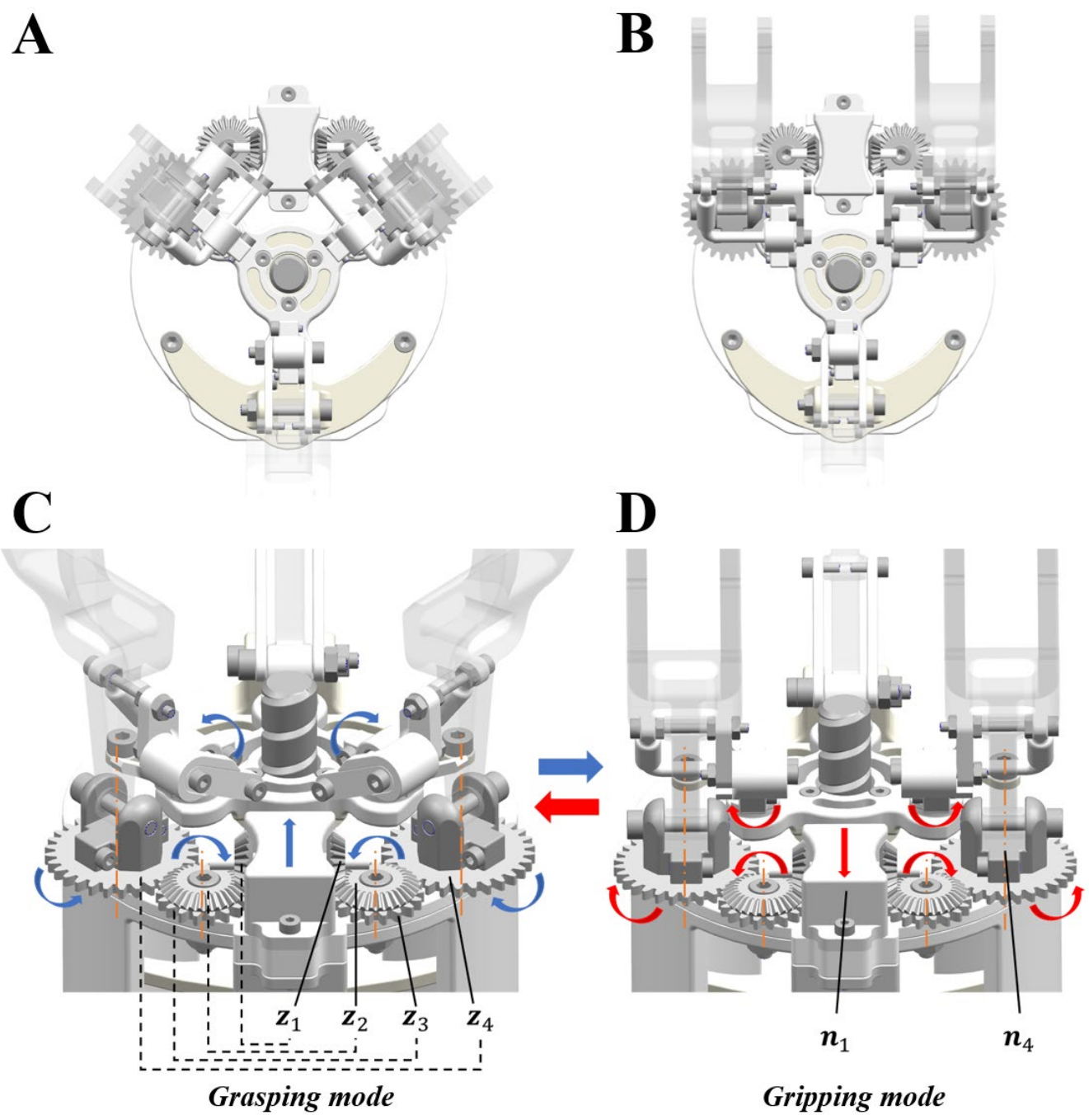

**Fig.3. Mode switching of the fractal gripper.** (**A** and **B**) Visualization along the screw axis for different modes of the fractal gripper is presented. (**C** and **D**) The mode switching process of the fractal gripper is shown, where (**C**) is the grasping mode and (**D**) is the gripping mode.

During the mode switching process, the worm-and-gear biaxial microform motor is utilized to drive the axle drive bevel pinion. This, in turn, engages the gear transmission to drive the floating gear seat, achieving the effect of bilateral anisotropic rotation. The floating gear seat then causes the rocker to rotate. The motion rod and its fittings within the cylindrical holes of the floating support seat allow for the relative motion of the rocker's middle section, designed to prevent interference during mode switching and stabilize the rocker's rotation. Additionally, the mode switching mechanism benefits from a self-locking effect, attributed to the worm-and-gear biaxial microform motor's reverse self-locking characteristic inherent in the worm-and-gear mechanism.

Mode switching is achieved with the use of a single actuator. Figure 3 illustrates the operation of the components during the transition between grasping and gripping modes, facilitated by the interaction between the motion rod and moving rod fitting to ensure a smooth changeover process. With three rockers in place, spatial grasping in different mode postures (grasping and gripping) is realized. The rocker design incorporates two rotatable and one non-rotatable component, determined by the interaction with the floating gear seat and the fixed support seat. The fixed support seat, in conjunction with the oscillating rod and the floating support seat, allows one rocker to achieve solely swinging effects. The other two rockers, through their reliance on the floating gear seat, the motion rod, its fittings, and the floating support seat, are enabled to perform both rotating and swinging actions.

The micro gear motor utilized for mode switching is of a dual-axis output type, featuring a symmetrical gear transmission structure with mirror symmetry on both sides. The motor's output shaft is matched with the active bevel gear, enabling the idler to drive through the active bevel gear's driving effect. The idler, comprising a combination of the driven bevel gear and the straight-tooth cylindrical gear, interacts with the floating gear holder to facilitate the rotation of the rocker. The active bevel gear is specified with $Z_1 = 20$, the driven bevel gear with $Z_2 = 30$, the straight-toothed cylindrical gear within the idler gear with $Z_3 = 18$, and the floating gear seat with $Z_4 = 30$. The loading velocity of the output shaft of the mode switching motor is given as $n_1 = 14$ r/min, and the velocity of the floating gear seat, denoted as $n_2$, can be derived based on the gearing ratio formula that relates the number of teeth to rotational velocity, see formula (3).

$$\frac{z_2 \cdot z_4}{z_1 \cdot z_3} = \frac{n_1}{n_4} \tag{3}$$

The rotational velocity of the floating gear seat is obtained as $n_4 = 5.6$ r/min, which contributes to the safety and stability of the mode switching process by keeping the velocity low. To adjust the mode switching velocity, adjustments need only be made to the number of teeth on different gears. However, in the process of adjusting the number of teeth, the load capacity of the transmission mechanism must be considered to ensure the system's stability and reliability. Consequently, the gears' number of teeth should not be set too small to ensure that the transmission mechanism possesses adequate load-bearing capacity.

Grasping mode is utilized in a broad spectrum of applications, ranging from fruit picking to vertical grasping. Through grasping mode, robots are enabled to efficiently acquire objects, such as hanging fruits or those located in elevated positions. Conversely, the gripping mode is found suitable for tasks involving horizontal movement, such as transferring a cylindrical object from one horizontal location to another. Different scenarios necessitate distinct grasping motions, underscoring the importance of targeting the grasping operation with various modal postures. Through the flexible selection of grasping attitudes, adaptation to a broad spectrum of object shapes and environmental conditions is enabled, facilitating the realization of accurate, efficient, and safe grasping tasks by the robot.

**Performance evaluation of the grasping space**

The design of the grasping space is primarily associated with the opening and closing angle of the fractal gripper and the length of the connecting rod. The opening and closing angle $\theta$ of the rocker is influenced by the lengths of different connecting rods. The travel of the movable nut and the length of the connecting rod constrain each other. Determine the appropriate size of the connecting rod by controlling the displacement of the movable nut and the swing limit position of the rocker with a travel of the movable nut of $L$ = 15 mm. The lengths of the oscillating rod and the motion rod have been determined to be 24.67 mm and 21.12 mm, respectively. During the mode switching process of the fractal gripper, the travel stroke of the motion rod is established as $s$ = 19 mm, and the floating support seat, together with the floating gear seat, is rotated through an angle of 50°. The spatial limit positions for opening and closing in different modes are depicted (Fig. 4).

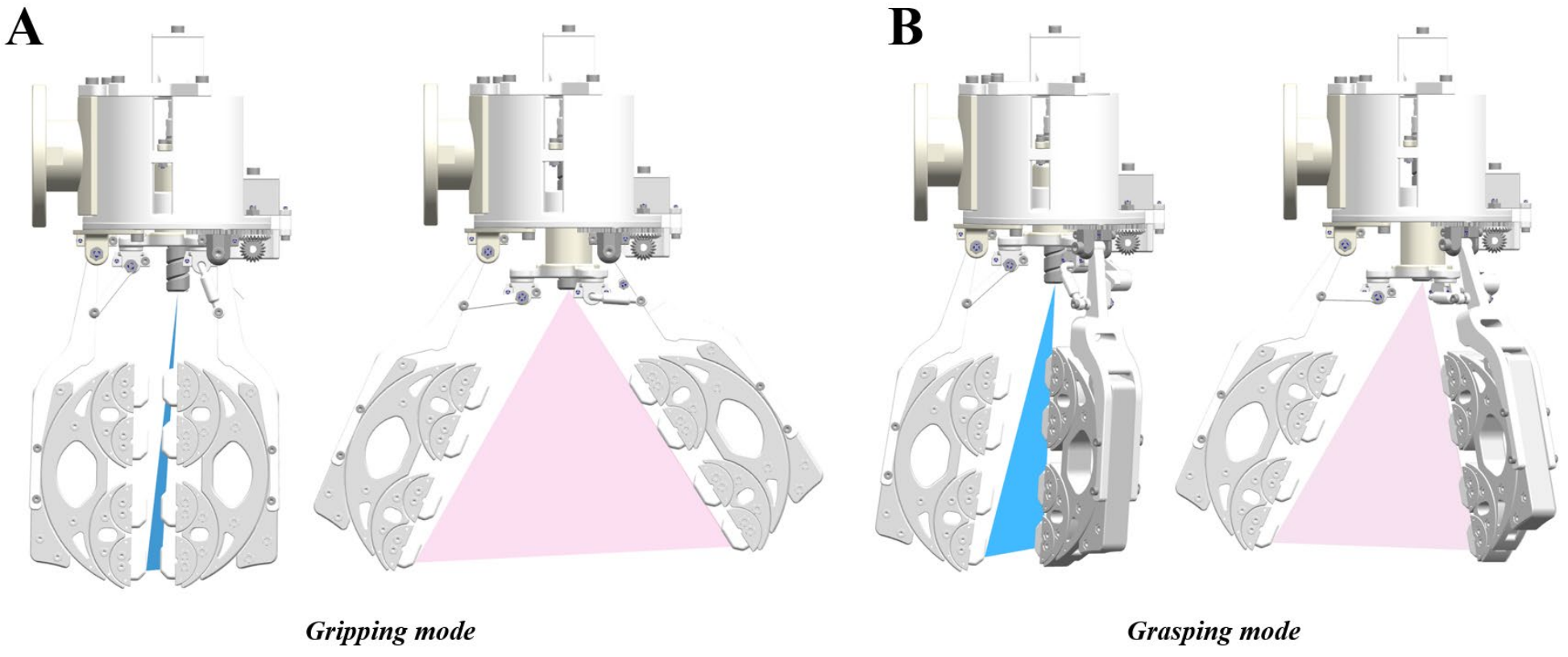

**Fig.4. The opening and closing space limit positions in different modes.** (**A**) Visualization of the four-point area volume in the Grasping mode. (**B**) Visualization of the four-point area volume in the Gripping mode.

An axonometric diagram of the limiting positions of the opening and closing spaces in different modes is presented (fig. S3). Meanwhile, an actualized prototype of the fractal gripper, demonstrating the limited positions of the opening and closing space in different modes, is depicted in (fig. S4). The measures for the limit position of opening and closing in different modes can be evaluated by employing a four-point area volume, with the state of the gripper closed in different modes being indicated by the blue area, and the state of the gripper open in different modes being indicated by the pink area in Fig. 4. The four-point region volume is constructed by selecting four calibration points: the center of the rotating screw's end face, and the midpoint of the boundary of the most anterior flexible body in the fractal structure. A comparison of the four-point region volume between the open and closed states can then be obtained, as shown in Table 2.

**Table 2. Four-point area volume comparison.**

| Four-point area volume ($cm^3$) | Open state | Close state |
|---|---|---|
| Grasping mode | 134.76 | 2.65 |
| Gripping mode | 147.51 | 6.99 |

With a travel of $L$=15mm for the movable nut, it is observed that the volume of the four-point area in the open state expands by 98.03% in the grasping mode and by 95.26% in the gripping mode, compared to the closed state.

The extent of the grasping space is indicated by the spatial limit positions of opening and closing, which are influenced by several factors. In the design of an improved grasping space, the compactness and strength of the overall structure are considered essential. To achieve a superior range of grasping space, the compactness of the overall structure must be ensured by rationalizing the arrangement and layout of the individual components to maximize the use of available space. Additionally, the strength of the grasping space, essential for withstanding the weight of the grasped object and external forces, is an important consideration. In the design process, it is necessary to balance the strength of the grasping space with its compactness to provide stable and reliable grasping capability. Therefore, achieving a superior grasping space design requires the coordination of the grasping space's strength with the structure's compactness, ensuring that the size requirements of the grasping space are effectively met while providing a reliable grasping function.

**Physical testing of the fractal gripper**

To showcase the superior grasping capability of the fractal gripper, a physical prototype was designed and constructed. The aim was to validate its performance through real-world spatial grasping tests. The gripper, predominantly 3D printed, utilized PLA engineering plastic and white resin to ensure adequate strength and reliability. The dimensions of the gripper were maximized to a length of 235.7 mm, a width of 151.8 mm, and a height of 215.2 mm, with an overall weight of 0.445 Kg. This approach emphasizes the practical application of the design and material selection in achieving desired outcomes.

In constructing the prototype, adherence to precise dimensional and geometric specifications was essential to ensure accuracy and stability. Subsequently, a series of spatial grasping tests were undertaken. These tests involved designing various scenarios and complex grasping tasks, during which the fractal gripper, suspended from a fixed bracket, utilized a wireless control module to enable the micro-motor's forward and reverse motions for handling real-world physical grasping tasks. The performance of the fractal gripper across diverse situations was visually assessed through the recording of physical tests. The outcomes of these physical grasping tests are illustrated (Fig. 5), demonstrating the fractal gripper's multi-scale adaptability capabilities.

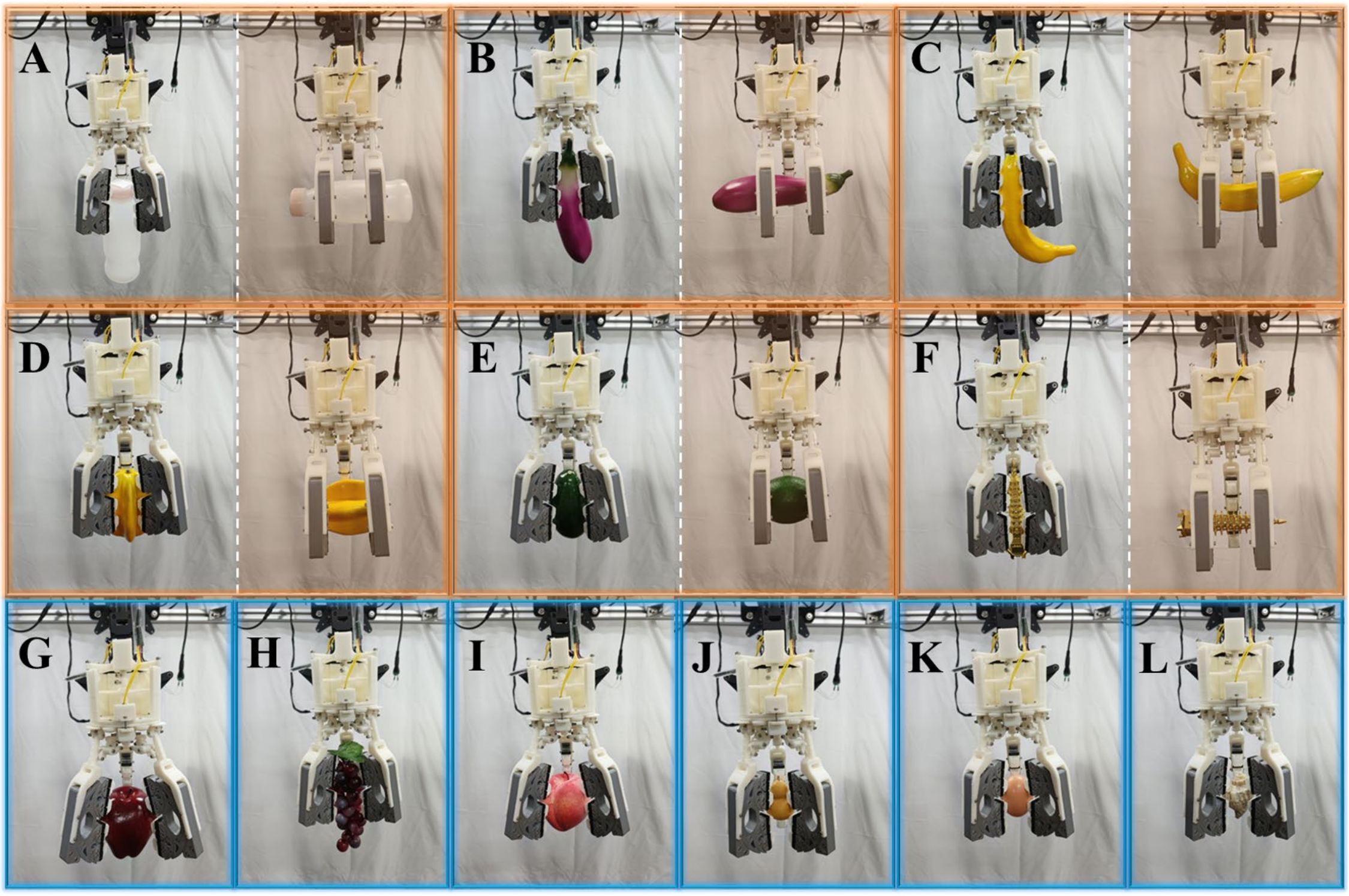

**Fig.5. Physical grab test of fractal gripper.** In the physical object grasping test, 12 different objects were grasped. The serial numbers **A-F** represent the adaptive grasping of the same object under different modes, including objects such as a bottle, eggplant, banana, starfruit, avocado, and pagoda. In the images of the gripping mode, the background is darkened for distinction from the grasping mode. Serial numbers **G-L** denote the remaining physical tests conducted in the grasping mode, involving objects like an apple, grapes, peach, gourd, egg, and conch, fully demonstrating the adaptive grasping capabilities of the fractal gripper in real-world scenarios.

The creation and grasping test of the physical prototype holds significant value in demonstrating the fractal gripper's exceptional grasping capability. These tests comprehensively showcase the gripper's excellent dual-mode versatility and adaptive characteristics, underscoring its practical effectiveness in various scenarios.

## 3 DISCUSSION

In this work, we designed a fractal gripper with mode switching. The gripper has been shown to possess good self-adaptability. During the object grasping process, reliance on three rockers to realize the passive wrapping of the fractal structure is observed, along with dependence on the self-locking characteristics of the screw drive, enabling effective grasping of the target object. The gripper was designed to be reproducible with a desktop-grade 3D printer, and the assembly of the fractal structure is facilitated by the half-and-half design of each fractal part. When the springs are installed, consideration of the specifications corresponding to each level is necessary. Beyond the gripper presented here, the miniature motor can be controlled remotely by a wireless control module, facilitating manual testing. Forward and reverse control of the motor is supported by the module, and it is also compatible with other control schemes.

We value fractal grippers for their reliable grasping performance, ease of integration (modularity), and simple control strategies, and encourage the industry to draw inspiration from the century-old fractal vise mechanism. Our goal is to see the creation of more exceptional passive fractal grippers that can serve as effective tools at the end of a robotic arm. Fractal gripper boasts features such as multi-scale adaptation (self-adaptive), force distribution characteristics, grasping stability (self-locking performance), versatility in grasping space (dual-mode switching), self-resetting capabilities, and durability against wear and tear. The integration of the fractal gripper at the robotic arm end simplifies the complexity found in the perception and grasping planning algorithms when compared to an anthropomorphic dexterous hand, significantly enhancing both the quality and efficiency of grasping. Drawing from the basic principle of the fractal vise, the unique adaptive characteristics of the fractal structure present in the original fractal vise were inherited, with the addition of a spring reset function to the foundational design, broadening its application in the task of grasping objects within three-dimensional spaces to forge an improved grasping mechanism. To safeguard the geometric surface of objects during the grasping process, a flexible polymer (PDMS) is applied to the wrapping surface of the fractal structure. Pressure is widely distributed over the target object's surface through the fractal structure of the gripper, and pressure is applied at multiple contact points. This effectively prevents deformation or damage to the object due to excessive localized stresses during closure. The force distribution characteristics of the fractal gripper can be clearly observed based on physical grasping experiments (fig. S5).

During the mode switching of the fractal gripper, the traveling stroke of the motion rod ranged from 0 to 19 mm, and the turning angle of the floating support seat alongside the floating gear seat varied from 0 to 50°. Given the traveling stroke of the movable nut $L$ = 15mm, it was observed that the volume of the four-point area in the open state of the gripping mode increased by 98.03% compared to the closed state, and similarly, the volume of the four-point area in the open state of the grasping mode saw an expansion of 95.26% compared to its closed state. This effectively showcases the variable degree of grasping space offered by the fractal gripper, which is capable of adaptively enveloping a diverse array of object shapes within the grasping space. As for additional applications of fractal grippers, their utility extends to underwater task retrieval, rehabilitation medical equipment, and heavy industry support equipment. The advancement of fractal gripper technology is anticipated to contribute to its widespread adoption in the future. Designed to deliver high-performance, safe, and reliable manipulators for the sectors of industry, agriculture, and robotics.

# 4 MATERIALS AND METHODS

## Printing of the fractal gripper

The overall fabrication of the fractal gripper is accomplished using only a 3D printer, alongside numerous nuts and screws. Rapid conversion from complex computer models to physical prototypes is enabled by utilizing 3D printing technology, significantly reducing the design-to-realization cycle. The flexibility of this technology allows for rapid iteration of designs, with design adjustments based on test feedback that is unmatched by traditional manufacturing techniques. Initially, materials such as PLA or ABS can be utilized for prototyping, offering not only cost-effectiveness but also a quick method to verify the validity of the fractal gripper design. With the progression of development, the introduction of high-performance materials, such as carbon fiber-reinforced plastics or metal alloys, for more demanding use conditions, not only enhances the strength and durability of the gripper but also optimizes its weight and efficiency. While performance is being enhanced, the availability of materials, the difficulty of processing, and cost-effectiveness must be considered by researchers to ensure the technology's broad applicability and economic viability. The overall weight of the assembled fractal gripper, including screws, nuts, springs, and the motor, has been determined by our team to be 0.445 Kg. PLA has been selected as the material, and the filling density for the different printed parts was subjectively chosen between 60% and 100%, ensuring the strength of the overall structure while considering weight minimization. Please refer to the supplementary materials attached to the article for further important details on the fractal gripper and fabrication considerations. Tables S1 to S3 provide details on the number of parts required for the construction of the fractal gripper and the STL file names for these parts.

**Addition of spring for self-resetting**

The self-resetting function of the fractal structure can be achieved by adding spring to the assembled structure, with only one end of the spring needing to be fixed. It is suggested that a cylindrical pin or wire be used at the outer end of the fractal structure to secure one end of the spring, setting the other end into a free state, and allowing some distance to be maintained from the bump in the middle of the arc groove. The spring should not be compressed too tightly. This prevents difficulties in adding another part of the spring and avoids increasing the gripper's clamping force and pressure on the object's surface. The length and diameter of the springs are selected based on the dimensions of the different arc slots. It is recommended that two springs of the same type be placed side by side in each arc slot. This arrangement helps to prevent the torsional entanglement of individual springs during the relative rotation of the fractal parts.

**Assembly of the mode switching mechanism**

Compared to the assembly of the fractal structure, the assembly of the mode switching part of the structure is considered to be a more difficult step. It is recommended that the active bevel gear be configured first on the output shaft of the worm-and-gear biaxial microform motor, using a transition fit. The torque from the active bevel gear is transmitted to the idler pulley, which then transmits it to the floating gear seat. This process, in cooperation with the floating support seat and the motion rod, realizes mode switching. The floating gear seat and floating support seat, both featuring a half-and-half design, are assembled on the circular bulge of the fixed plate using screws, enabling circumferential rotation. During assembly, the floating gear seat is first assembled onto the fixed plate, followed by the assembly of the idler pulley, and finally, the worm-and-gear biaxial microform motor, with attention being paid to the fit between the gears and the mirror image arrangement of the floating gear seat on both sides.

# APPENDIX

We provide the additional details for the fractal gripper in the Appendix.

**The Appendix includes:**

Figs S1 to S5

Tables S1 to S3

**Table S1.** Summary of 3D printing files for fractal structures

| File Name | Model Display* | Required Quantity |
|---|---|---|
| Top of the fractal rack.stl | 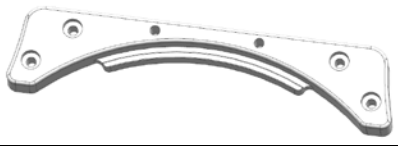 | 1 |
| Bottom of the fractal rack.stl | 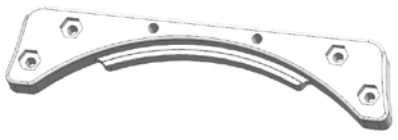 | 1 |
| Fractal first piece top.stl | 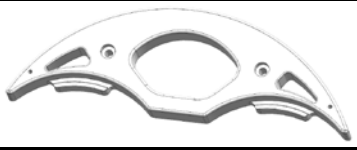 | 1 |
| Fractal first piece bottom.stl | 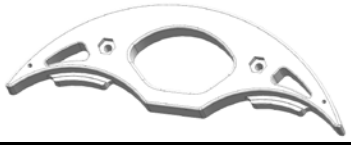 | 1 |
| Fractal second piece top.stl | 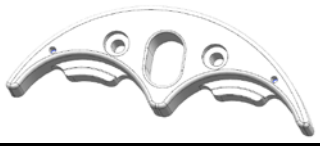 | 2 |
| Fractal second piece bottom.stl | 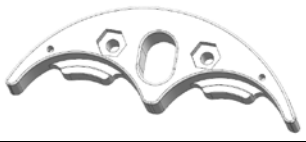 | 2 |
| Fractal third piece top.stl | 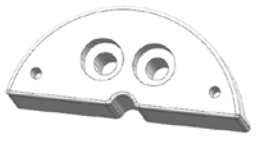 | 4 |
| Fractal third piece bottom.stl | 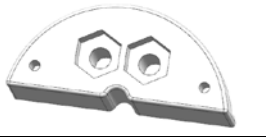 | 4 |

*Note: the scale of the Model Display is not to actual proportion.

**Table S2.** Summary of 3D printed files for gripper (without fractal structure)

| File Name | Model Display* | Required Quantity |
|---|---|---|
| Axle drive bevel pinion.stl | 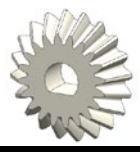 | 2 |
| Coupling.stl | 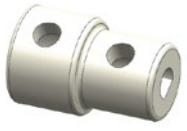 | 1 |
| Fixed pad.stl | 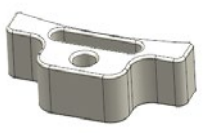 | 1 |
| Fixed support seat.stl | 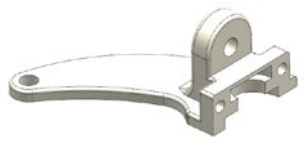 | 1 |
| Fixed support seat_mirror image.stl | 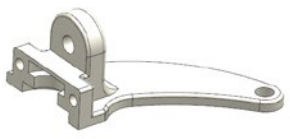 | 1 |
| Flange fixing block.stl | 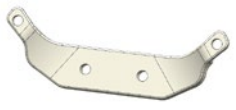 | 1 |
| Flange.stl | 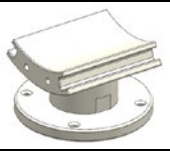 | 1 |
| Floating gear seat.stl | 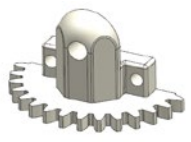 | 4 |
| Floating support seat.stl | 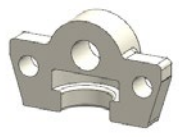 | 6 |
| Idler.stl | 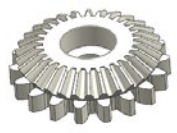 | 2 |
| Motion rod.stl | 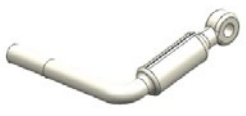 | 2 |
| Motor bracket fittings.stl | 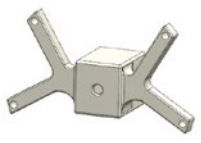 | 1 |
| Motor bracket.stl | 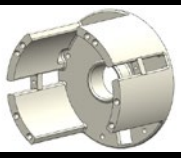 | 1 |
| Motor cover fitting.stl | 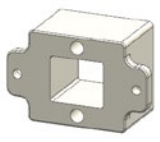 | 1 |
| Motor cover plate.stl | 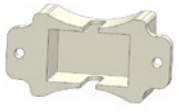 | 1 |
| Motor fixing plate.stl | 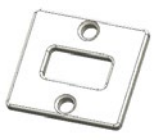 | 2 |
| Movable nut fitting.stl | 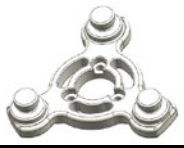 | 1 |
| Movable nut.stl | 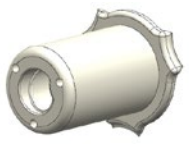 | 1 |
| Moving rod fitting_1.stl | 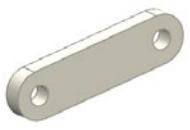 | 2 |
| Moving rod fitting_2.stl | 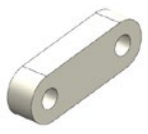 | 2 |
| Oscillating rod.stl | 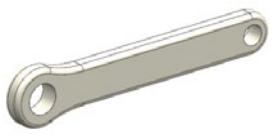 | 2 |
| Rocker.stl | 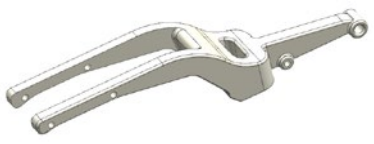 | 3 |
| Rotating screw.stl | 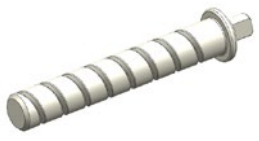 | 1 |
| Screw support.stl | 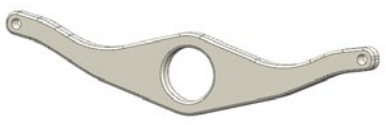 | 1 |
| Sleeve limit.stl | 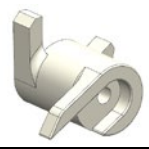 | 1 |
| Sleeve limit_mirror image.stl | 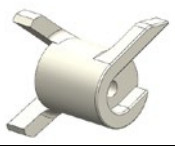 | 1 |
| Support plate.stl | 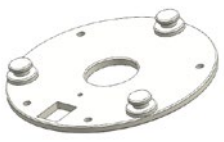 | 1 |

*Note: the scale of the Model Display is not to actual proportion.

**Table S3.** Summary of all parts for assembling fractal gripper

| Name of parts | Required Quantity |
|---|---|
| Fractal Structure | 3 |
| Gripper | 1 |
| Micro gear motor | 1 |
| Biaxial microform motor | 1 |
| Helical compression spring | numerous |
| Screw and nut | numerous |

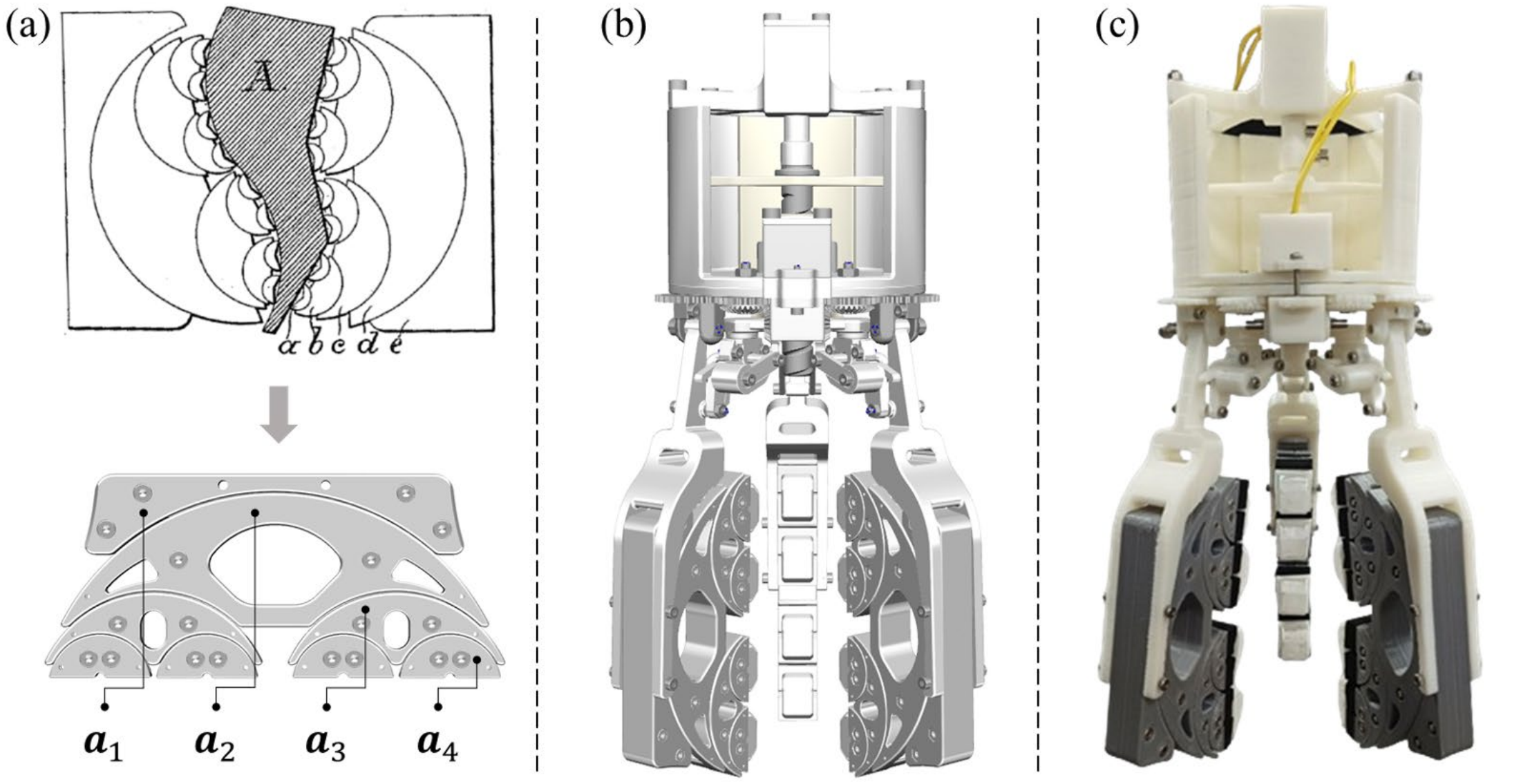

**Figure S1**. Sources of fractal structures and physical references to the fractal gripper. (a) The design of the fractal structure is informed by Kunze's patent (*49*), and the components shown in the original patent include the fingers *α*, and the connecting layers *b* through *e*. These components work in concert to form a complex and robust structure for clamping and securing object *A*. In the fractal structure described in this work, there are four main parts: the fractal rack ($a_1$), the fractal first piece ($a_2$), the fractal second piece ($a_3$), and the fractal third piece ($a_4$). The connection between each part is achieved using T-slots. (b) Demonstrates model visualization of the 3D design of a fractal gripper. (c) Demonstrates the visualization of a fractal gripper from a 3D model to a physical object, with a scale size of 1:1. All STL files for the fractal gripper are shown in Table S1 and Table S2.

## The production details of the fractal gripper

To illustrate some of the fabrication details of the fractal gripper, a description is provided below in conjunction with Figure S2. The names of the files constituting the fractal structure are listed in Table S1. The STL files for the fractal structure have been made available, allowing for reproduction through the 3D printing process. A demonstration of the slicing effect for the 3D printed file is presented in Figure S2, where printing with a fill density of 90% or higher is recommended when using PLA or other plastic materials.

The flexible body is attached to the fractal third piece ($a_4$) within the fractal structure, with a description provided of the production process for fabricating the corresponding PDMS structure. First, a cavity container needs to be prepared; then, the PDMS liquid is poured into the container and cured with temperature; finally, the cured flexible body is obtained. In the liquid pouring step for PDMS, the main agent and curing agent must be mixed in a 1:1 ratio by weight or volume. After thorough mixing, the product's thickness should be observed for either atmospheric pressure or vacuum degassing. It is recommended that the curing temperature be increased to quickly shorten the curing time. However, for thicker products, caution is advised as too rapid curing may result in the formation of air bubbles.

During the assembly of the fractal gripper, it should be noted that some parts feature a half-and-half design. These parts (e.g., the floating gear seat) must be assembled to achieve a clearance fit with the circular bulge, allowing it to rotate circumferentially after assembly; otherwise, mode switching of the fractal gripper will not be feasible. For each part with a half-and-half design, through holes are reserved in the structure, which facilitates the use of screws and nuts for fixing.

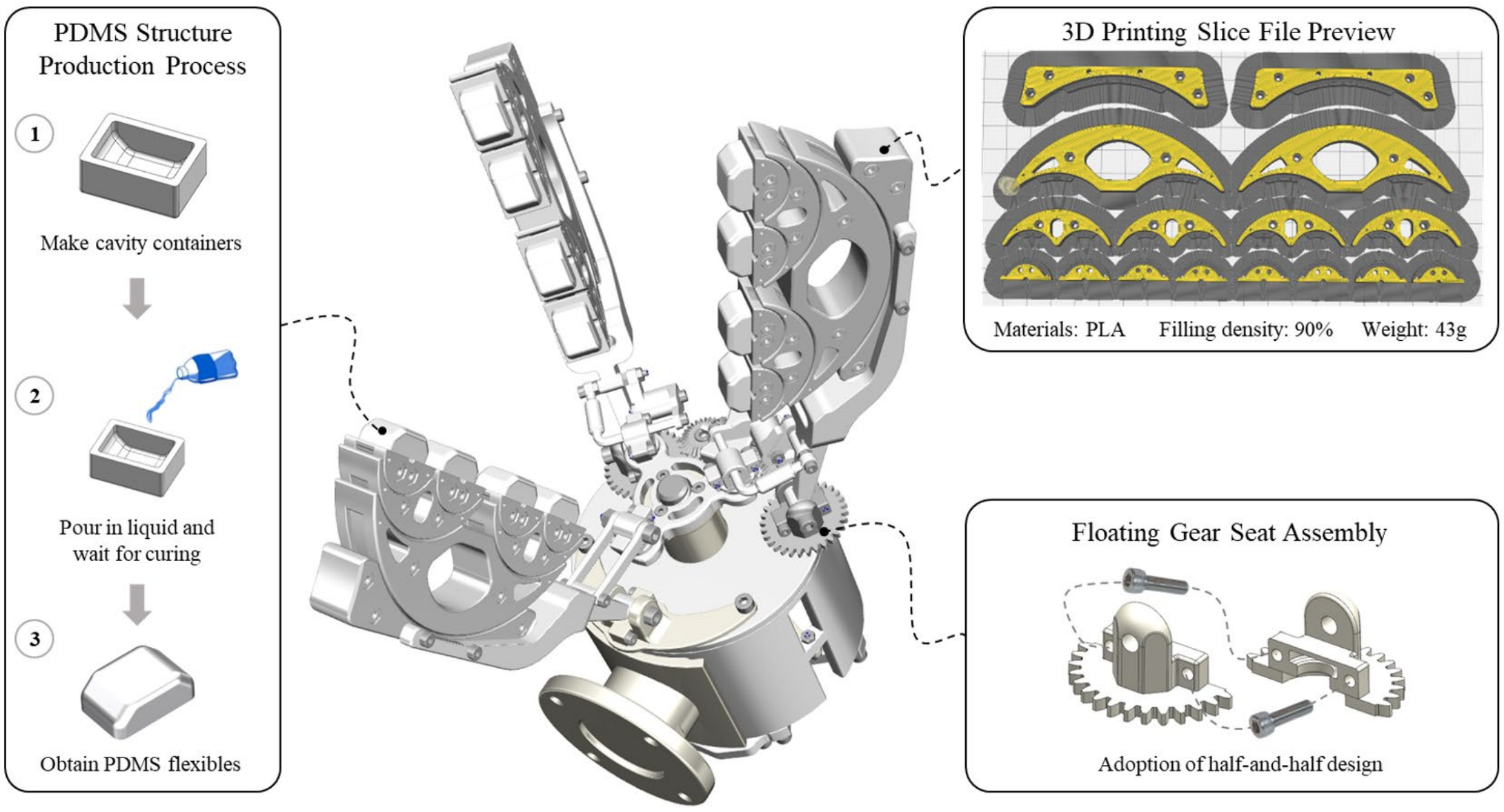

**Figure S2.** Partial details regarding the fabrication and assembly of the fractal gripper are presented.

## Grasping space and physical demonstration of the fractal gripper

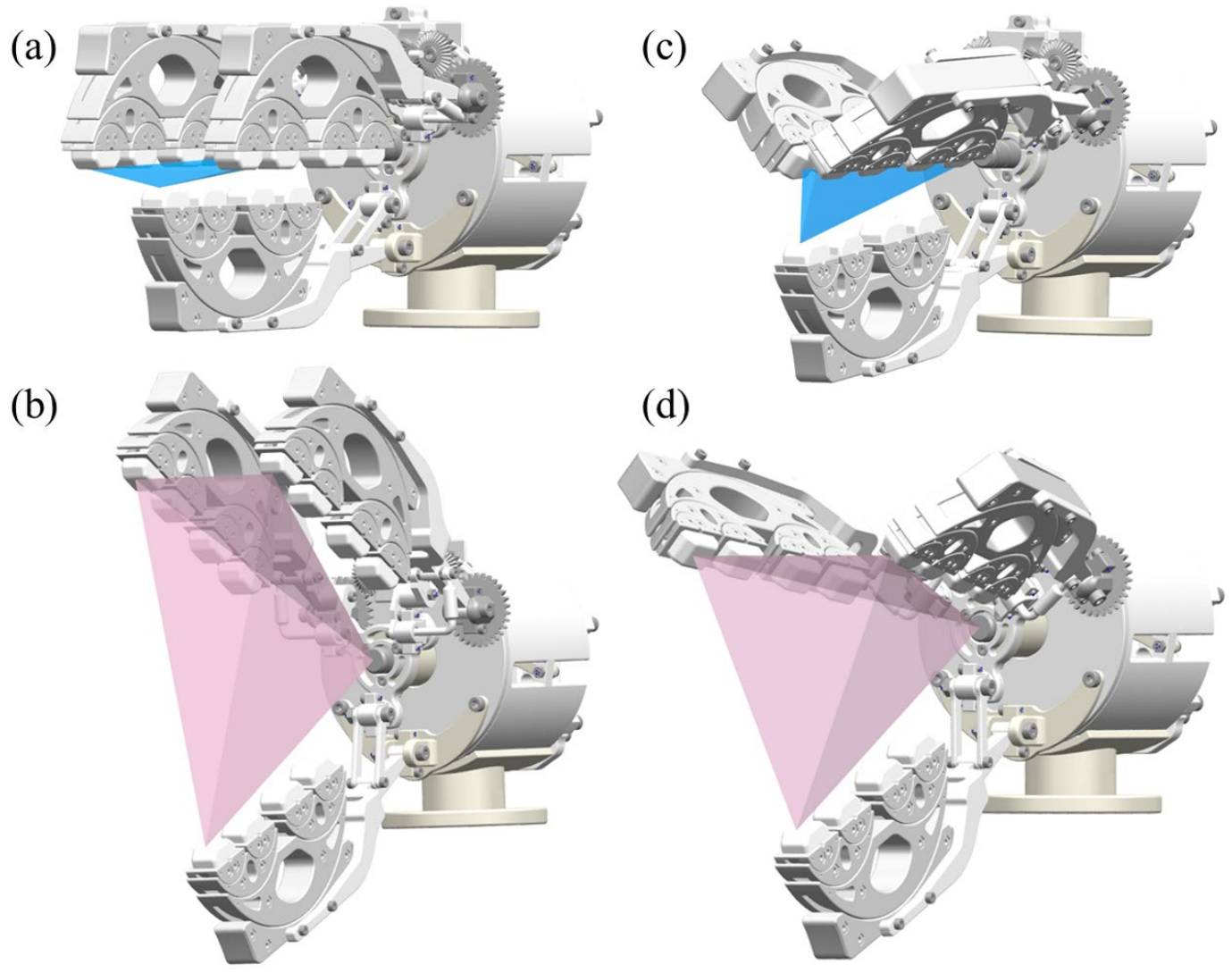

**Figure S3.** Axonometric diagram of opening and closing space limit positions in different modes. The blue and pink areas in the figure represent the visualization of the volume of the four-point region. (a) represents the volume visualization of the four-point region in the closed state of the gripping mode. (b) represents the volume visualization of the four-point region in the open state of the gripping mode. (c) represents the volume visualization of the four-point region in the closed state of the grasping mode. (d) represents the volume visualization of the four-point region in the open state of the grasping mode.

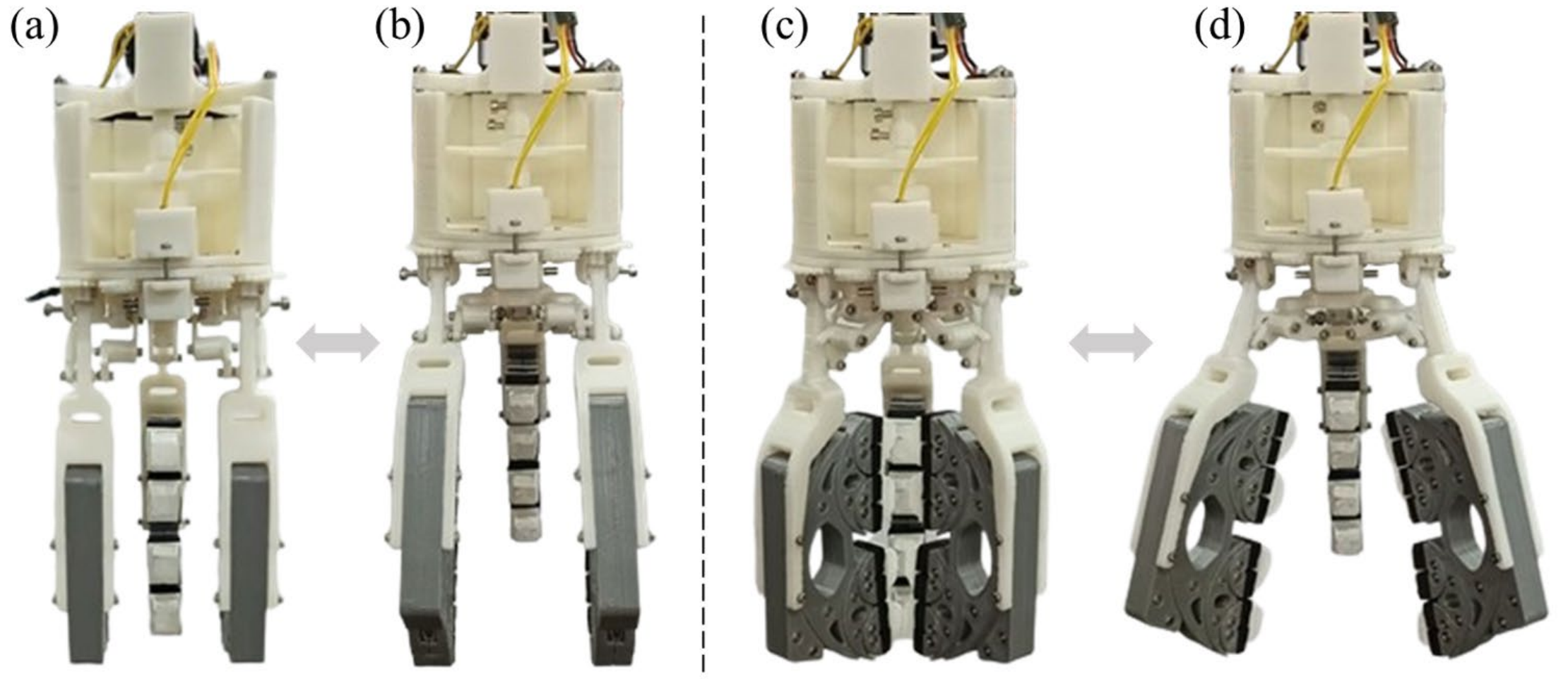

**Figure S4.** The mode switching, along with the opening and closing actions of the fractal gripper, is physically demonstrated.

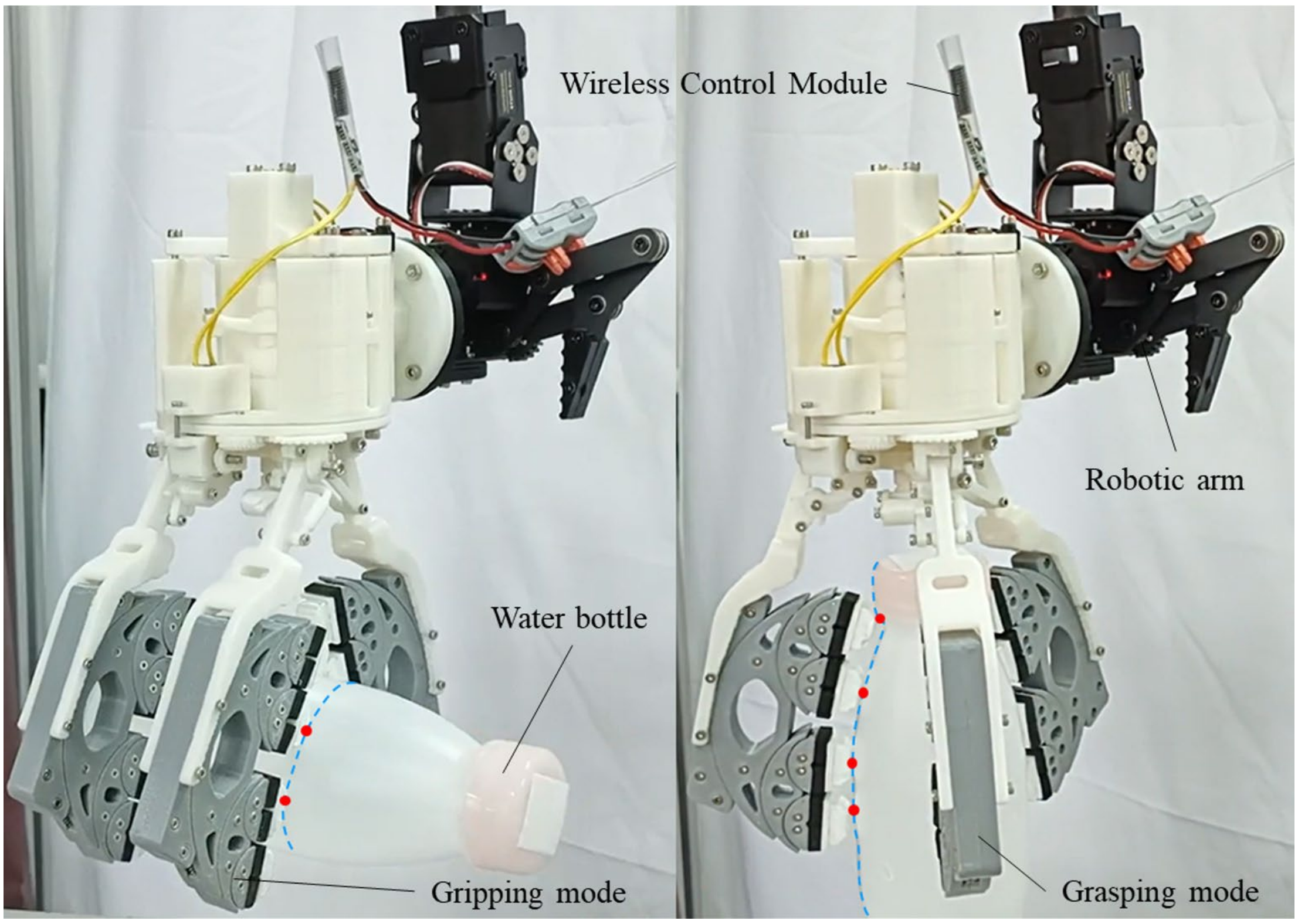

**Figure S5.** Physical grab demo of a fractal gripper, analyzing a bottle as a case study. The fractal gripper is attached to the end position of the robotic arm by a flange. A demonstration was conducted to grasp the same bottle using different modes, and in combination with the experimental picture, the fractal gripper's grasp stability was observed to be excellent. In the gripping mode, two flexible bodies in each fractal gripper come into contact with the surface of the bottle, resulting in a force distribution effect. In the grasping mode, four flexible bodies in each fractal gripper come into contact with the surface of the bottle, resulting in a more pronounced force distribution effect. In the figure, the contour of the bottle is depicted by the blue dotted line, while the red dot indicates the point of contact between the flexible body and the bottle's surface contour.